\documentclass{article} 
\usepackage{iclr2027_conference,times}

\usepackage{amsmath,amsfonts,bm}

\def\eqref#1{equation~\ref{#1}}

\def\1{\bm{1}}

\DeclareMathAlphabet{\mathsfit}{\encodingdefault}{\sfdefault}{m}{sl}
\SetMathAlphabet{\mathsfit}{bold}{\encodingdefault}{\sfdefault}{bx}{n}

\newcommand{\Nmethods}{7}
\newcommand{\Nratio}{3}
\newcommand{\Nseeds}{3}
\newcommand{\Ntotal}{93}
\newcommand{\nLarge}{18}
\newcommand{\CompHours}{18.8}
\newcommand{\Nsteps}{900}
\newcommand{\Nlr}{$1{\times}10^{-3}$}
\newcommand{\ProbeEvery}{50}
\newcommand{\PfMax}{0.400}
\newcommand{\EvalAt}{225/450/675/900}
\newcommand{\Ld}{320}
\newcommand{\Llayers}{8}
\newcommand{\Nconfigs}{63}
\newcommand{\NckptsAll}{372}
\newcommand{\rhoCosglobal}{-0.321}
\newcommand{\rhoCosglobalLo}{-0.523}
\newcommand{\rhoCosglobalHi}{-0.097}

\newcommand{\rhoConflictrate}{0.338}
\newcommand{\rhoConflictrateLo}{0.131}
\newcommand{\rhoConflictrateHi}{0.537}

\newcommand{\rhoConflictmag}{-0.009}
\newcommand{\rhoConflictmagLo}{-0.266}
\newcommand{\rhoConflictmagHi}{0.268}

\newcommand{\rhoNormratio}{-0.293}
\newcommand{\rhoNormratioLo}{-0.507}
\newcommand{\rhoNormratioHi}{-0.035}

\newcommand{\rhoRemovedenergy}{0.366}
\newcommand{\rhoRemovedenergyLo}{0.159}
\newcommand{\rhoRemovedenergyHi}{0.559}

\newcommand{\rhoCoswdepth}{-0.295}
\newcommand{\rhoCoswdepthLo}{-0.535}
\newcommand{\rhoCoswdepthHi}{-0.040}

\newcommand{\rhoCoswshallow}{-0.171}
\newcommand{\rhoCoswshallowLo}{-0.428}
\newcommand{\rhoCoswshallowHi}{0.083}

\newcommand{\Nckpts}{252}

\newcommand{\nEarly}{63}

\newcommand{\rhoCkCosglobalPartial}{-0.147}
\newcommand{\rhoCkCosglobalPartialP}{0.019}

\newcommand{\rhoCkConflictratePartial}{0.124}
\newcommand{\rhoCkConflictratePartialP}{0.049}

\newcommand{\rhoCkConflictmagPartial}{-0.192}

\newcommand{\rhoCkNormratioPartial}{-0.528}

\newcommand{\rhoCkRemovedenergyPartial}{0.178}

\newcommand{\rhoLossPareto}{-0.926}
\newcommand{\rhoLossParetoLo}{-0.959}
\newcommand{\rhoLossParetoHi}{-0.883}
\newcommand{\rhoLossParetoP}{0.000}
\newcommand{\domPairs}{90}
\newcommand{\domTotal}{1953}
\newcommand{\domFrac}{0.046}
\newcommand{\rhoLossU}{-0.921}
\newcommand{\rhoLossG}{-0.835}
\newcommand{\rhoLossEarly}{-0.424}
\newcommand{\rhoLossEarlyLo}{-0.623}
\newcommand{\rhoLossEarlyHi}{-0.173}

\newcommand{\DoseMagZero}{0.005}
\newcommand{\DoseMagOne}{0.000}
\newcommand{\rhoDoseMan}{1.000}
\newcommand{\rhoDoseCaus}{-0.600}
\newcommand{\rhoDoseCausP}{0.285}
\newcommand{\nDose}{12}

\newcommand{\rhoPredCosglobal}{-0.146}
\newcommand{\rhoPredCosglobalLo}{-0.411}
\newcommand{\rhoPredCosglobalHi}{0.147}
\newcommand{\rhoPredConflictrate}{0.022}
\newcommand{\rhoPredConflictrateLo}{-0.210}
\newcommand{\rhoPredConflictrateHi}{0.260}
\newcommand{\rhoPredConflictmag}{-0.198}
\newcommand{\rhoPredConflictmagLo}{-0.463}
\newcommand{\rhoPredConflictmagHi}{0.079}
\newcommand{\rhoPredNormratio}{-0.006}
\newcommand{\rhoPredNormratioLo}{-0.264}
\newcommand{\rhoPredNormratioHi}{0.246}
\newcommand{\rhoPredRemovedenergy}{0.052}
\newcommand{\rhoPredRemovedenergyLo}{-0.185}
\newcommand{\rhoPredRemovedenergyHi}{0.287}
\newcommand{\predEffrank}{0.702}
\newcommand{\predEffrankLo}{0.603}
\newcommand{\predEffrankHi}{0.782}
\newcommand{\predNormratio}{-0.651}
\newcommand{\predNormratioLo}{-0.721}
\newcommand{\predNormratioHi}{-0.573}
\newcommand{\predProbeacc}{0.534}
\newcommand{\predProbeaccLo}{0.406}
\newcommand{\predProbeaccHi}{0.639}
\newcommand{\predConflictmag}{-0.187}
\newcommand{\predConflictmagLo}{-0.318}
\newcommand{\predConflictmagHi}{-0.046}
\newcommand{\predRemovedenergy}{0.179}
\newcommand{\predRemovedenergyLo}{0.062}
\newcommand{\predRemovedenergyHi}{0.294}
\newcommand{\predSubspaceoverlap}{0.132}
\newcommand{\predSubspaceoverlapLo}{-0.037}
\newcommand{\predSubspaceoverlapHi}{0.281}
\newcommand{\predConflictrate}{0.123}
\newcommand{\predConflictrateLo}{0.007}
\newcommand{\predConflictrateHi}{0.239}
\newcommand{\predSelfconsistency}{-0.117}
\newcommand{\predSelfconsistencyLo}{-0.259}
\newcommand{\predSelfconsistencyHi}{0.034}
\newcommand{\predCosglobal}{-0.074}
\newcommand{\predCosglobalLo}{-0.190}
\newcommand{\predCosglobalHi}{0.045}
\newcommand{\bestAlt}{eff\_rank}

\newcommand{\predBestalt}{0.702}
\newcommand{\predBestaltLo}{0.603}
\newcommand{\predBestaltHi}{0.782}

\newcommand{\bestConfDir}{conflict\_mag}
\newcommand{\predBestconfDir}{-0.187}

\newcommand{\deltaAltConfDirLo}{0.729}
\newcommand{\deltaAltConfDirHi}{1.041}
\newcommand{\absGapAltNr}{0.051}
\newcommand{\absGapAltNrLo}{-0.044}
\newcommand{\absGapAltNrHi}{0.140}

\newcommand{\IccAccu}{0.982}
\newcommand{\IccAccgObj}{1.000}
\newcommand{\reliabCeiling}{0.991}
\newcommand{\rhoGlobalDisatt}{-0.324}
\newcommand{\mdeConfig}{0.362}
\newcommand{\mdeCkpt}{0.178}
\newcommand{\mCagradU}{0.636}

\newcommand{\mCagradObj}{0.999}

\newcommand{\mCagradPareto}{0.818}
\newcommand{\mCagradCos}{-0.003}
\newcommand{\mCagradRate}{0.708}

\newcommand{\mCagradNratio}{0.029}
\newcommand{\mGradnormU}{0.633}

\newcommand{\mGradnormObj}{0.996}

\newcommand{\mGradnormPareto}{0.815}
\newcommand{\mGradnormCos}{0.002}
\newcommand{\mGradnormRate}{0.514}

\newcommand{\mGradnormNratio}{0.112}
\newcommand{\mMgdaU}{0.389}

\newcommand{\mMgdaObj}{1.000}

\newcommand{\mMgdaPareto}{0.694}
\newcommand{\mMgdaCos}{0.123}
\newcommand{\mMgdaRate}{0.208}

\newcommand{\mMgdaNratio}{0.004}
\newcommand{\mNaiveU}{0.630}

\newcommand{\mNaiveObj}{0.998}

\newcommand{\mNaivePareto}{0.814}
\newcommand{\mNaiveCos}{0.001}
\newcommand{\mNaiveRate}{0.417}

\newcommand{\mNaiveNratio}{0.025}
\newcommand{\mPcgradU}{0.641}

\newcommand{\mPcgradObj}{0.999}

\newcommand{\mPcgradPareto}{0.820}
\newcommand{\mPcgradCos}{0.000}
\newcommand{\mPcgradRate}{0.639}

\newcommand{\mPcgradNratio}{0.027}
\newcommand{\mSoapU}{0.543}

\newcommand{\mSoapObj}{0.000}

\newcommand{\mSoapPareto}{0.271}
\newcommand{\mSoapCos}{0.002}
\newcommand{\mSoapRate}{0.389}

\newcommand{\mSoapNratio}{0.476}
\newcommand{\mToknormU}{0.622}

\newcommand{\mToknormObj}{0.607}

\newcommand{\mToknormPareto}{0.614}
\newcommand{\mToknormCos}{0.000}
\newcommand{\mToknormRate}{0.542}

\newcommand{\mToknormNratio}{0.339}
\newcommand{\frontierN}{3}
\newcommand{\frontierMethods}{pcgrad (2), cagrad (1)}

\newcommand{\bestAltHealthy}{self\_consistency}
\newcommand{\predBestaltHealthy}{0.854}

\newcommand{\predBestconfHealthy}{0.230}
\newcommand{\deltaAltConfHealthyLo}{0.470}
\newcommand{\deltaAltConfHealthyHi}{0.784}

\newcommand{\absMaxDir}{0.145}

\newcommand{\ConcConfigClaim}{(concurrent configuration level: cos global $\rho = -0.32$ [-0.52, -0.10]; conflict rate $\rho = +0.34$ [0.13, 0.54]; removed energy $\rho = +0.37$ [0.16, 0.56]---moderate, mutually sign-inconsistent, and unstable under the regime cut below)}

\newcommand{\NormRatioClaim}{The exception is the norm ratio, a magnitude-imbalance quantity rather than a directional conflict measure---and its mechanism is failure detection, not interference: across the full checkpoint population it reaches $\rho = -0.65$ (95\% CI [-0.72, -0.57]), but it correlates with generation accuracy at $\rho = -0.80$, and within the configurations that master generation---the regime in which the field would actually deploy it---it is null ($\rho = +0.02$, CI [-0.17, 0.18])}
\newcommand{\HealthyClaim}{Repeating the checkpoint-level analysis within the 135 checkpoints of configurations that master generation sharpens the picture: every directional conflict metric stays weak and sign-inconsistent (largest $|\rho| = 0.23$), the norm ratio drops to null ($\rho = +0.02$), and the functional measures stay strong (probe separability $\rho = +0.69$, effective rank $\rho = +0.47$)}
\newcommand{\ConfClaimWtd}{(with the shallow-weighted variant the exception: $\rho = -0.29$, CI [-0.55, -0.01], below the $|\rho| \geq 0.3$ threshold)}
\newcommand{\LayerClaim}{an otherwise flat profile with block 5 the exception ($\rho = -0.36$, CI [-0.59, -0.11])}

\newcommand{\LossClaim}{a strong configuration-level surrogate in this testbed ($\rho = -0.92$ within the 47 generation-mastering configurations), although 74 of 1081 loss-ordered pairs still reverse}

\newcommand{\DoseClaim}{flat within seed noise (total range 0.005 against a mean per-dose s.d. of 0.028)}
\newcommand{\alphaDelta}{-0.002}
\newcommand{\alphaDeltaU}{-0.004}
\newcommand{\alphaDeltaG}{0.001}
\newcommand{\alphaSpread}{0.037}
\newcommand{\AlphaClaim}{moves the joint score by -0.002 (understanding -0.004, generation object-cell +0.001), a shift within the seed-to-seed spread (0.037) of the un-intervened configuration}
\newcommand{\ScaleClaim}{the directional conflict metrics remain within noise at the larger scale (largest $|\rho| = 0.15$, CI [-0.33, 0.04], against 0.19 at the base scale), and the norm-ratio association persists ($\rho = -0.42$, CI [-0.64, -0.20])}

\newcommand{\BestClaim}{large and statistically significant ($|\rho|$ gap 0.51)}
\newcommand{\ConclusionVerdict}{no more than weak and statistically fragile}
\newcommand{\DiscussionStrength}{weak at the predictive level (largest $|\rho| = 0.14$, interval crossing zero) and unstable where it is larger (the concurrent configuration-level associations are sign-inconsistent and dissolve under the regime cut)}

\usepackage{hyperref}
\usepackage{url}
\usepackage{xcolor}
\usepackage{booktabs}
\usepackage{amsmath,amssymb}
\usepackage{graphicx}
\usepackage{multirow}
\usepackage{tikz}
\usetikzlibrary{arrows.meta,positioning,fit,decorations.pathmorphing}

\newcommand{\gridumm}{\textsc{GridUMM}}
\newcommand{\accu}{\mathrm{acc}_u}
\newcommand{\accg}{\mathrm{acc}_{g}}
\newcommand{\Nblocks}{6}
\newcommand{\Ngroupsn}{8}

\title{Does Gradient Conflict Predict the \\Understanding--Generation Trade-off? \\A Controlled Audit of Conflict-Metric Validity in Unified Multimodal Models}

\author{%
	\begin{tabular}{@{}p{0.45\textwidth}@{\hspace{0.06\textwidth}}p{0.45\textwidth}@{}}
		\begin{minipage}[t]{\linewidth}
			\vspace{0pt}
			\raggedright
			\normalfont
			\textbf{Shuyang Jiang}\\
			University of California, Los Angeles\\
			\texttt{shuyangjiang@ucla.edu}
		\end{minipage}
		&
		\begin{minipage}[t]{\linewidth}
			\vspace{0pt}
			\raggedright
			\normalfont
			\textbf{Fucheng Deng}\\
			Aimakj\\
			\texttt{dengfucheng@aimakj.com}
		\end{minipage}
		\\[3em]
		\begin{minipage}[t]{\linewidth}
			\vspace{0pt}
			\raggedright
			\normalfont
			\textbf{Yuchuan Luo}\\
			College of Computer Science and Technology\\
			National University of Defense Technology\\
			\texttt{luoyuchuan09@nudt.edu.cn}
		\end{minipage}
		&
		\begin{minipage}[t]{\linewidth}
			\vspace{0pt}
			\raggedright
			\normalfont
			\textbf{Zhenyu Wu}\thanks{Corresponding author.}\\
			Key Laboratory of Advanced\\
			Microprocessor Chips and Systems\\
			College of Computer Science and Technology\\
			National University of Defense Technology\\
			\texttt{wuzhenyu@nudt.edu.cn}
		\end{minipage}
	\end{tabular}%
}

\iclrfinalcopy

\begin{document}

\maketitle

\begin{abstract}
Unified multimodal models (UMMs) are increasingly designed and optimized
around the notion of \emph{gradient conflict} between the understanding and
generation objectives, measured as cosine similarities between task gradients
or conflict rates across layers. The premise that reducing these metrics
improves the downstream understanding--generation trade-off has never been
tested directly. We audit it in a controlled testbed, \gridumm{}, which
mirrors the structural ingredients of UMM training---a fully shared trunk
serving two objectives with asymmetric token budgets and task
difficulty---while making the ground-truth trade-off exactly computable.
Across \Nconfigs{} configurations (\Nmethods{} gradient-combination
strategies $\times$ \Nratio{} data-mixing ratios $\times$ \Nseeds{} seeds)
and \NckptsAll{} measured checkpoints, no directional conflict
metric---global or per-layer cosine, conflict rate, removed energy---reaches
$|\rho| \geq 0.3$ with a confidence interval excluding zero for conflict
measured during training against the eventual trade-off (strongest
$|\rho| = \absMaxDir{}$, interval crossing zero), and the larger concurrent
configuration-level associations are mutually sign-inconsistent. A
dose--response intervention that monotonically suppresses conflict (level
$\alpha$) leaves the trade-off flat, separating correlation from causation.
The one gradient-geometric quantity with association, the norm ratio, is a
generation-failure detector, null among configurations that master
generation. The functional interference measure \bestAlt{} leads every
directional conflict metric ($\Delta\rho$ CI
\deltaAltConfDirLo{}--\deltaAltConfDirHi{}) and retains its advantage inside
that regime. Training loss, the outcome most often reported in place of
benchmarks, tracks the trade-off strongly---the failure is specific to
gradient-conflict geometry. Our results do not show that conflict is
useless; they show that its validity as a diagnostic target must be
established, not assumed, and we release the audit protocol as a reusable
standard.
\end{abstract}

\section{Introduction}
\label{sec:intro}

Unified multimodal models (UMMs) such as Show-o \citep{showo}, Emu3
\citep{emu3}, Janus and Janus-Pro \citep{wu2025janus,januspro2025},
Chameleon \citep{chameleon2024}, Transfusion \citep{transfusion2024}, and
Unified-IO~2 \citep{unifiedio2} serve visual understanding and visual
generation from a single transformer, and a growing family of work motivates
interventions by pointing to \emph{gradient conflict}: during joint
training, the understanding and generation objectives produce gradients whose
cosine similarity is near zero or negative, with magnitude ratios that can
exceed an order of magnitude \citep{rao2026dofight}. Uni-X
\citep{hao2026unix} separates conflicting shallow and deep layers
architecturally; Symbiotic-MoE \citep{liu2026symbioticmoe} shields
conflict-driven forgetting with expert disentanglement; ML-FOP-SOAP
\citep{lu2026mlfopsoap} suppresses modality conflicts with Fisher-orthogonal
projection and second-order preconditioning; Pareto LoRA
\citep{paretolora2026} rebalances modality gradients by Pareto-optimal
integration. Across this literature one premise recurs untested:
\emph{that lowering a gradient-conflict metric moves the model toward a
better understanding--generation trade-off}.

The premise deserves scrutiny because its two ends are measured on
different scales: conflict metrics describe gradient geometry on a fixed
probe batch, while the trade-off is a property of two benchmark families
evaluated after training under an evaluation practice that treats the two
objectives independently \citep{wang2026crosstask,unieval2025}. Existing
evidence is consistent with decoupling: across seven training strategies and
two post-hoc methods on Janus-Pro, none improved generation CLIPScore
($p>0.5$) even though all modify conflict \citep{rao2026dofight};
\citet{lu2026mlfopsoap} report training-loss gains without downstream
evaluation; and high understanding and high generation scores do not imply
alignment \citep{wang2026crosstask}. But no study has measured the quantity
that would settle the premise---the correlation between conflict metrics and
trade-off position across a controlled population of training runs---because
on production-scale UMMs this requires dozens of full training runs.

This paper performs that audit in a controlled testbed, organized around
three design requirements. First, the ground-truth trade-off must be
computable exactly, ruling out benchmark noise from dominating a correlation
analysis: we construct \gridumm{}, a synthetic environment whose two tasks
have exactly measurable accuracies and whose Pareto frontier over
configurations can be computed without approximation \citep[cf.][]{deb2002},
while reproducing the structural ingredients of UMM training---a fully shared
trunk, two asymmetric objectives, an order-of-magnitude token-budget
asymmetry, and autoregressive generation against single-token
classification---so that the conflict phenomenology matches what has been
reported on production UMMs \citep{rao2026dofight}. Second, the audit must
include an intervention, not only an observational correlation: we suppress
conflict by projecting the generative gradient with controlled strength
$\alpha$, moving the conflict level monotonically while holding data,
schedule, and parameters fixed. Third, the audit must be honest about
measurement noise: we quantify the test--retest reliability of every
downstream metric and report the ceiling that any correlation cannot exceed
\citep{spearman1904,shrout1979}.

Our contributions are as follows.
\begin{itemize}
\item \textbf{An audit framework} combining observational correlations with
bootstrap intervals, partial correlations controlling for training progress,
reliability ceilings, and a causal dose--response intervention; the field's
implicit premise becomes four falsifiable hypotheses, each with a killing
experiment.
\item \textbf{A controlled testbed}, \gridumm{}, reproducing the
gradient-conflict phenomenology reported on production UMMs at a compute
scale that permits \Nconfigs{} configurations; we release the world, the
protocol, and all measurement code.
\item \textbf{The audit result}: the directional conflict metrics---global
and per-layer cosines, conflict rate, removed energy---are weak, fragile
predictors of the trade-off, and suppressing conflict does not improve it;
the norm ratio, the one gradient-geometric quantity with association, is a
generation-failure detector. Training loss tracks the trade-off strongly,
locating the failure in gradient-conflict geometry specifically, while
functional interference measures predict it better than every conflict
metric and retain that advantage in the regime that masters generation.
\item \textbf{A diagnostic standard}: a protocol any UMM practitioner can
run to decide whether conflict reduction is a valid intermediate target in
their setting, and what to measure instead.
\end{itemize}

A controlled testbed cannot establish that any particular
production-scale conflict-metric claim is false; it establishes what a
validity claim must demonstrate, shows that the demonstration is absent from
the literature, and makes such a demonstration routine.
Section~\ref{sec:discussion} returns to the implications for current
conflict-motivated UMM designs.

\section{Related Work}
\label{sec:related}

\paragraph{Unified architectures.}
A first family of UMMs trains one transformer over interleaved discrete
tokens for both objectives: Show-o unifies autoregressive and
discrete-diffusion modeling \citep{showo}, Emu3 demonstrates pure next-token
prediction over interleaved tokens \citep{emu3}, and Janus, Janus-Pro,
Chameleon, and Transfusion establish decoupled visual encoding, early-fusion
token mixing, and combined language-modeling plus diffusion objectives
\citep{wu2025janus,januspro2025,chameleon2024,transfusion2024}; later
systems add progressive vocabularies, continuous tokens, visual
autoregression, instruction tuning, and native unified generation
\citep{ugen2025,unifluid2025,vargpt2025,showo2,metamorph2025,hunyuanimage3,
pisces2025} \citep[cf.][]{ummsurvey2025}. Our testbed abstracts the shared
structure common to these designs---one trunk, one parameter space, two
objectives with unequal token budgets
(Appendix~\ref{sec:app:related})---while replacing natural data with a
grammar whose ground-truth trade-off is exactly computable.

\paragraph{Conflict-motivated UMM design and diagnosis without validation.}
A family of 2025--2026 works takes gradient conflict between the two
objectives as the design driver: Uni-X separates conflicting shallow and
deep layers \citep{hao2026unix}, Symbiotic-MoE shields conflict-driven
forgetting \citep{liu2026symbioticmoe}, ML-FOP-SOAP suppresses
variance-induced conflicts with Fisher-orthogonal projection and SOAP-style
preconditioning \citep{lu2026mlfopsoap,vyas2024soap}, Pareto LoRA integrates
imbalanced modality gradients \citep{paretolora2026}, and task-aware MoE
routing resolves task-objective conflicts \citep{taskmoe2025}. For each, the
conflict metric is a premise and the trade-off is reported separately. Closest
to questioning the premise, \citet{rao2026dofight} measure near-orthogonal
gradients and find no DPO variant improves generation CLIPScore, and
cross-task consistency benchmarks \citep{wang2026crosstask,unieval2025} show
that the two scores do not imply alignment; synergy, adversary, and
continual-training studies characterize the phenomenon or mitigate its
symptoms \citep{wu2026synergy,su2026unigame,intrainter2025}. None validates
the diagnostic metric itself.

\paragraph{Multi-task optimization.}
The optimizer-side interventions borrowed by UMM work originate in
multi-task learning \citep{caruana1997,mtlsurvey2021}, with shared
architectures from routing \citep{mmoe2018} to task-interaction networks
\citep{mtinet2020}, loss weights adapted by uncertainty
\citep{kendall2018} or auxiliary-gradient magnitudes \citep{metabalance2022},
and gradient manipulation reshaping geometry directly: minimum-norm
combinations \citep{sener2018mgda}, gradient surgery and conflict-averse
updates \citep[see][]{mtlsurvey2021,liu2021cagrad}, bargaining-game
solutions \citep{nashmtl2022}, independent-component alignment
\citep{senushkin2023}, with Pareto concepts from multi-objective
optimization \citep{deb2002}. Our audit uses faithful implementations of
these methods as the population over which conflict metrics vary---if
conflict metrics fail to predict trade-offs even across methods whose entire
purpose is to reshape gradient geometry, the metric's diagnostic value is in
doubt.

\paragraph{Modality imbalance.}
Joint multimodal training is known to favor the faster-learning modality:
the greedy-dynamics analysis of \citet{peng2020hard} explains why multimodal
networks can underperform unimodal ones, and balancing methods modulate
gradients or prototypes on the fly \citep{ogmge2022,pmr2023,otfm2024,
towardseq2025,scope2026}. Our norm-ratio finding connects to this literature:
magnitude imbalance is real and diagnostic, but of generation failure, not of
a directional conflict signal (Section~\ref{sec:e1}).

\paragraph{Audits of benchmark validity.}
Our protocol follows the tradition of audits that re-examine whether an
accepted evaluation practice measures what it claims: rebuilt test sets
revealed overfitting to reused ImageNet splits \citep{recht2019,
vasudevan2022}, controlled re-evaluation collapsed metric-learning gains
\citep{musgrave2020}, and methodological reviews document the cost of
unvalidated proxies \citep{varoquaux2022}. To our knowledge, no analogous
audit exists for training-time conflict metrics in UMMs; this paper supplies
one. Appendix~\ref{sec:app:related} discusses each of these five
literatures in more detail.

\section{A Controlled Testbed for Conflict-Metric Validity}
\label{sec:testbed}

\subsection{Why a controlled testbed}
\label{sec:why}
Auditing the validity of a diagnostic metric requires varying the metric
while holding everything else fixed, measuring the ground-truth target
without noise, and intervening on the metric causally. On a production UMM
the correlation analysis alone would require $\geq$\Nconfigs{} full training
runs per design point, and the trade-off is observed through benchmarks
whose seed-to-seed variance can exceed the differences being measured
(Section~\ref{sec:reliab}). A controlled environment solves both problems at
once and makes the audit pre-registerable; the price is external validity,
managed by matching the structural ingredients of UMM training and verifying
that the testbed's conflict phenomenology reproduces production measurements
(Section~\ref{sec:experiments}).

\gridumm{} places two objectives over one decoder-only transformer. The
data world is a $5\times 5$ grid over eight colors; a scene contains 6--8
non-overlapping rectangular objects, and both tasks share a single
interleaved marker/color token representation whose learned position markers
anchor each output slot to its coordinate. \emph{Generation} maps a
description---objects enumerated in randomized order---to the marker/color
stream autoregressively (25 color tokens plus end-of-stream, against 33--61
description tokens); \emph{understanding} answers a question about the same
stream (cell color, color count, or color-pair adjacency) with a
single-token answer. Held-out sets are exact samples of the same grammar
under fixed seeds, so downstream accuracies are unbiased estimates of
generalization. Two regime facts matter for reading the audit: generation
is mastered well before the budget ends by every configuration that trains
its gradient at natural scale---the one weighting scheme that scales it down
by the token-count ratio (token-normalized) fails to master it in a subset
of seeds, an imbalance pathology visible in Table~\ref{tab:methods}---so
most trade-off variation lives on the understanding axis; and the conflict
phenomenology's magnitude is concentrated in the learning phase, exactly
where the field's premise operates (Figure~\ref{fig:dynamics}, Appendix~\ref{sec:app:audit}). Every
parameter is shared by both tasks: the strongest form of the UMM design
point and the one under which conflict claims are made.

\begin{figure}[t]
\centering
\includegraphics[width=0.8\linewidth]{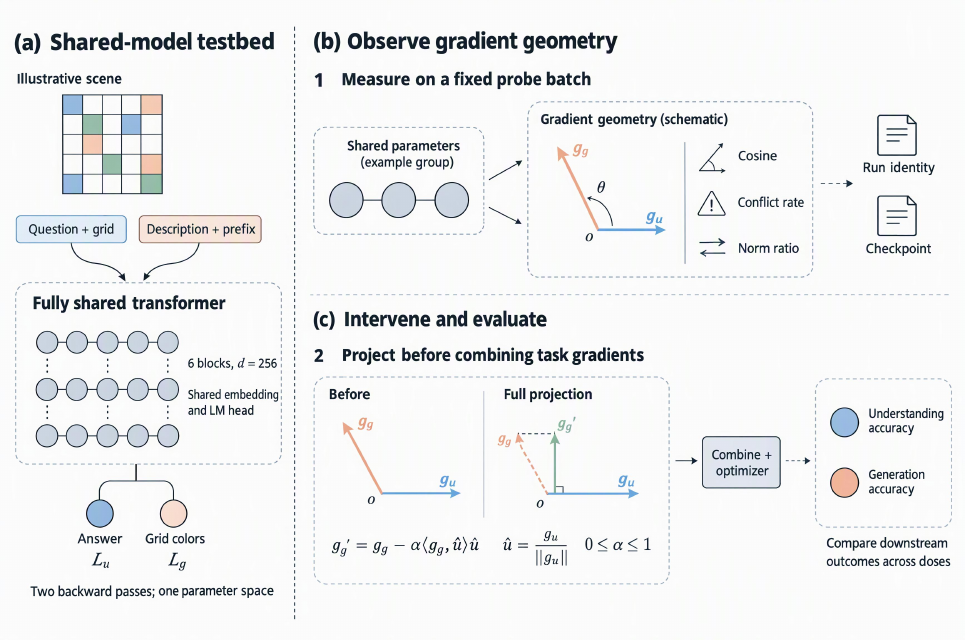}
\caption{\gridumm{} audit protocol. (a) Two task streams use the same
embedding, transformer and language-model head; the colored terminal nodes
denote outputs, not task-specific parameter branches. The grid and network
glyphs are schematic. (b) A fixed probe batch yields task gradients grouped
by shared parameters; geometry is matched to run identity and checkpoint.
(c) In E3 only, the $\alpha$-projection removes a component parallel to
$g_u$ before combination. Full projection illustrates $\alpha=1$;
it does not preserve the gradient norm. Held-out understanding and
generation scores evaluate the resulting model, without selecting updates.
No experimental data are depicted in this conceptual illustration.}
\label{fig:framework}
\end{figure}

\subsection{Training protocol}
\label{sec:protocol}
All runs use a \textbf{two-backward protocol}: each step draws a batch of
64 samples with the configuration's data-mixing ratio, computes
$\mathcal{L}_u$ and $\mathcal{L}_g$ in separate forward/backward passes, and
captures per-group gradients (embedding, each of \Nblocks{} blocks, and the
output head) per objective. The update direction is the combination
produced by the configuration's method (Appendix~\ref{sec:app:methods});
conflict statistics are measured on a \emph{fixed probe batch} (32+32
samples, identical across all runs) every 50 steps, so cross-configuration
comparisons are not confounded by batch composition.

Because the generation objective is autoregressive, we train it with a
\textbf{teacher/student-forcing mixture}: before step 200 the loss is pure
teacher forcing \citep{williamszipser1989} on the ground-truth prefix; the
student-forced share then ramps linearly to \PfMax{}, i.e., for that share
of samples the loss is computed on the model's own sampled prefix while the
targets remain ground truth. The mixture keeps the generation gradient
alive throughout training; Appendix~\ref{sec:app:proto} documents the
schedule. Training runs \Nsteps{} steps with AdamW (\Nlr{}, warmup 100,
cosine decay), batch 64, and the understanding-to-generation data ratio
drawn from $\{1{:}2, 1{:}1, 2{:}1\}$.

\subsection{Conflict metrics}
\label{sec:metrics}
At every snapshot we compute, over the shared parameters,
\begin{align}
\cos(g_u, g_g) \;&\;\text{(global and per-group)},\qquad
r \;=\; \lVert g_g\rVert_2 / \lVert g_u\rVert_2,\\
\text{conflict rate} \;&=\; \text{fraction of groups with } \cos_l < 0,\qquad
\text{conflict magnitude} \;=\; \tfrac{1}{K}\textstyle\sum_l \max(0, -\cos_l),
\end{align}
and the \emph{removed energy}, the fraction of $\lVert g_u\rVert^2$ a
gradient-surgery-style projection would remove, aggregated over groups; $K$
indexes the \Ngroupsn{} parameter groups. The norm ratio is reported in the
direction generation/understanding; in \gridumm{} the generation side has
the larger token budget (26 target tokens versus one answer token) but the
smaller gradient norm at convergence, reproducing the imbalance structure
reported on production UMMs \citep{rao2026dofight}.

\subsection{Downstream trade-off}
\label{sec:downstream}
Each run is evaluated at four checkpoints (\Nsteps{}/4 through \Nsteps{})
on a fixed held-out set: understanding accuracy $\accu$ over 512 questions
and generation $\accg$, the object-cell accuracy of greedy free-run
decoding over 256 held-out descriptions. To compare configurations on a
single axis we define the \textbf{Pareto score} $S = (\accu + \accg)/2$;
the two-dimensional plane and its empirical Pareto frontier are analyzed
directly as well (Figure~\ref{fig:pareto}). A configuration is
\emph{Pareto dominated} by another if the latter matches or exceeds it on
both axes with at least one strict inequality. We avoid min--max
normalization: with a saturating generation axis it would amplify resampling
noise into the ranking (Appendix~\ref{sec:app:proto}).

\subsection{Functional interference measures}
\label{sec:func}
Because our hypotheses contrast parameter-space gradient geometry with
\emph{functional} interference, we measure four candidates at each
checkpoint, drawing on the probing and representation-similarity literature
\citep{belinkov2022,klabunde2025}. (i)~\emph{Probe separability}: accuracy
of a linear probe on the trunk's mean-pooled last-block features predicting
the answer token of held-out questions (2-fold cross-validated).
(ii)~\emph{Subspace overlap}: mean principal-angle cosine between the top-8
singular subspaces of the two tasks' last-block feature matrices.
(iii)~\emph{Effective rank}: participation ratio of the covariance spectrum
of the pooled feature matrix. (iv)~\emph{Self-consistency}: the fraction of
questions about the model's \emph{own} generated scenes that it answers
correctly---a functional analogue of the cross-task consistency XTC-Bench
measures at the benchmark level \citep{wang2026crosstask}.

\subsection{Hypotheses}
\label{sec:hyps}
The premise decomposes into four falsifiable hypotheses, each with a
designated experiment. \textbf{HA-1} (the premise): some conflict metric
correlates strongly ($|\rho|\geq 0.3$) with the Pareto score across a
controlled population; killed by E1, a correlation sweep with bootstrap
intervals over \Nconfigs{} configurations and \Nckpts{} checkpoint rows,
with partial correlations controlling for training progress.
\textbf{HA-2}: training loss is a valid surrogate; tested by E2, the rank
correlation of final joint loss with Pareto score plus its predictive form
and a dominance analysis. \textbf{HA-3}: if HA-1 fails, functional measures
predict better; established or refuted by E4, a predictor competition with
paired bootstrap tests. \textbf{HA-4} (the adversarial converse, stated
before measurement to prevent post-hoc rationalization): conflict metrics
are predictive once appropriately weighted---by depth or stage; tested
inside E1 with depth-weighted cosines and stage-conditional analyses. E3
supplies causal evidence: if conflict is monotonically suppressed by an
intervention that changes nothing else and the trade-off does not respond
monotonically, no reweighting rescue can be purely causal. E5 replicates
the E1 sign structure at a second model scale.

\section{Experiments}
\label{sec:experiments}

\subsection{Setup}
\label{sec:setup}
The population consists of \Nconfigs{} base-scale configurations---\Nmethods{}
gradient-combination strategies $\times$ \Nratio{} data ratios
(1:2, 1:1, 2:1) $\times$ \Nseeds{} seeds---plus \nDose{} intervention
configurations (E3) and \nLarge{} configurations at a second scale (E5):
\Ntotal{} training runs, \NckptsAll{} measured checkpoints, each with the
full conflict-metric and downstream vectors (Table~\ref{tab:setup},
Appendix~\ref{sec:app:audit}). The
compute budget was a single Apple M4 Max laptop (MPS, float32), up to six
concurrent configurations, \CompHours{} aggregate run-hours; every manifest
records device, software, and wall-clock time. Methods are faithful
implementations of the published algorithms
(Appendix~\ref{sec:app:methods}); two documented simplifications are noted
there.

\begin{table}[t]
\centering
\caption{Main results by method (base scale, final checkpoint; mean $\pm$
s.d.\ over ratios and seeds, $n{=}9$ per method). Object-cell accuracy is
the generation axis; the joint score is their mean.}
\label{tab:methods}
\resizebox{0.96\linewidth}{!}{%
\begin{tabular}{lcccccc}
\toprule
Method & understanding & generation (obj.) & joint score &
$\cos(g_u,g_g)$ & rate & norm ratio \\
\midrule
Naive & \mNaiveU{} & \mNaiveObj{} & \mNaivePareto{} & \mNaiveCos{} &
\mNaiveRate{} & \mNaiveNratio{} \\
Token-normalized & \mToknormU{} & \mToknormObj{} & \mToknormPareto{} &
\mToknormCos{} & \mToknormRate{} & \mToknormNratio{} \\
GradNorm-style & \mGradnormU{} & \mGradnormObj{} & \mGradnormPareto{} &
\mGradnormCos{} & \mGradnormRate{} & \mGradnormNratio{} \\
PCGrad-style & \mPcgradU{} & \mPcgradObj{} & \mPcgradPareto{} & \mPcgradCos{} &
\mPcgradRate{} & \mPcgradNratio{} \\
CAGrad & \mCagradU{} & \mCagradObj{} & \mCagradPareto{} & \mCagradCos{} &
\mCagradRate{} & \mCagradNratio{} \\
MGDA & \mMgdaU{} & \mMgdaObj{} & \mMgdaPareto{} & \mMgdaCos{} &
\mMgdaRate{} & \mMgdaNratio{} \\
SOAP-style & \mSoapU{} & \mSoapObj{} & \mSoapPareto{} & \mSoapCos{} &
\mSoapRate{} & \mSoapNratio{} \\
\bottomrule
\end{tabular}}
\end{table}

\subsection{E1: does conflict predict the trade-off?}
\label{sec:e1}
The operative premise is predictive: practitioners measure conflict
\emph{during} training and read it as a signal about the eventual
trade-off. We test both readings. The \textbf{predictive} analysis
correlates conflict statistics measured at 25\% of the budget with the final
Pareto score ($N=\nEarly{}$); the \textbf{concurrent} analysis correlates
conflict and trade-off at the same checkpoint across all \Nckpts{} rows
with cluster-bootstrap intervals, and a partial Spearman correlation removes
the training-progress component (rank-regressing out the step index),
addressing the trivial concern that both quantities simply improve with
training.

\begin{figure}[t]
\centering
\includegraphics[width=0.7\linewidth]{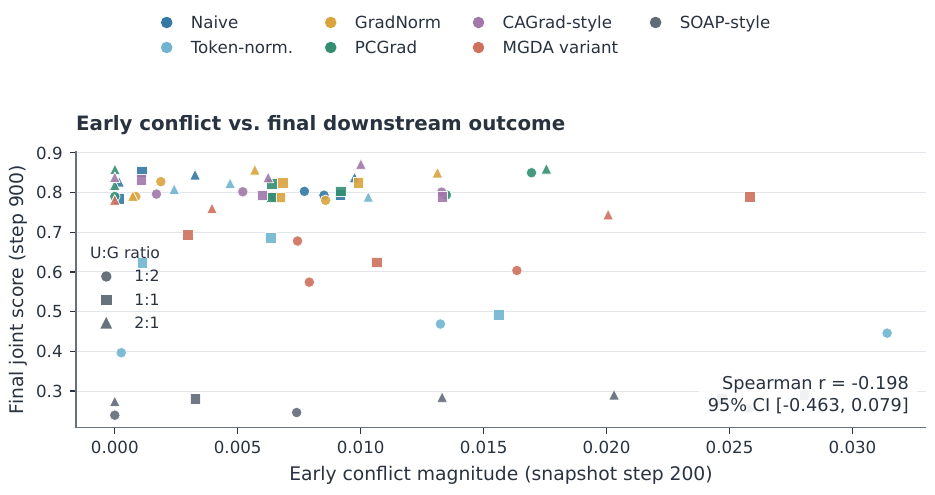}
\caption{The core audit: probe conflict at step 200 (associated with the
step-225 evaluation checkpoint) against the final joint score at step 900,
for all \Nconfigs{} configurations
(colors: methods; shapes: data ratios). The rank correlation is
$\rho = \rhoPredConflictmag{}$
[\rhoPredConflictmagLo{}, \rhoPredConflictmagHi{}].}
\label{fig:scatter}
\end{figure}

Figure~\ref{fig:scatter} shows the core result: conflict magnitude measured
during the learning phase against the final trade-off, every configuration
plotted. The rank correlation is $\rho = \rhoPredConflictmag{}$ (95\% CI
[\rhoPredConflictmagLo{}, \rhoPredConflictmagHi{}]). Table~\ref{tab:cor}
reports the full conflict-metric vector at both levels
(Appendix~\ref{sec:app:audit}): at the predictive level---the field's operative
reading---no directional conflict metric reaches $|\rho| \geq 0.3$ with a
confidence interval excluding zero. The early-concurrent HA-4 weighted
analysis gives \ConfClaimWtd{}; the per-layer profile
(Figure~\ref{fig:layer}) shows \LayerClaim{}. \NormRatioClaim{}. The
concurrent configuration-level associations are larger but mutually
sign-inconsistent \ConcConfigClaim{}. \HealthyClaim{}. At the concurrent
checkpoint level the partial correlations controlling for progress are
$\rho = \rhoCkCosglobalPartial{}$ (global cosine) and
$\rho = \rhoCkConflictratePartial{}$ (conflict rate), with
$p = \rhoCkCosglobalPartialP{}$ and \rhoCkConflictratePartialP{}
respectively.

\subsection{E2: is training loss a valid surrogate?}
\label{sec:e2}
Training loss is the outcome most often reported when benchmarks are
skipped---the practice our audit tests, following \citet{lu2026mlfopsoap}.
At the configuration level, the rank correlation of the final joint probe
loss with the Pareto score is $\rho = \rhoLossPareto{}$ (95\% CI
\rhoLossParetoLo{}--\rhoLossParetoHi{}, $p = \rhoLossParetoP{}$);
task-resolved correlations are $\rho = \rhoLossU{}$ (understanding) and
$\rho = \rhoLossG{}$ (generation). The predictive form---joint loss at
25\% of the budget read as a forecast of the eventual trade-off, which is
what a loss curve is used for---correlates at $\rho = \rhoLossEarly{}$
(95\% CI \rhoLossEarlyLo{}--\rhoLossEarlyHi{}). The run-level pairing
(Figure~\ref{fig:e2}) shows \domPairs{} of \domTotal{} loss-ordered
pairs (\domFrac{}) with the better-loss configuration Pareto-dominated. A
lower training loss is therefore \LossClaim{}. The contrast with E1 is
itself informative: in the same testbed where every directional conflict
metric fails, the loss surrogate tracks the trade-off, so the invalidity
exposed by this audit is specific to gradient-conflict geometry, not a
generic consequence of using cheap proxies.

\subsection{E3: intervention---suppressing conflict causally}
\label{sec:e3}
Correlation, however weak, cannot separate ``conflict is uninformative''
from ``conflict is informative but confounded.'' We therefore intervene: at
every step, the generative gradient is projected against the understanding
gradient with strength $\alpha$, scaling every group's task-gradient cosine
by $(1-\alpha)$ and hence monotonically suppressing conflict while holding
data, schedule, and parameters fixed. The manipulation check
(Figure~\ref{fig:e3}, left) confirms the monotone dose: as the learning-phase
mean over snapshots at steps 200--500, the effective conflict magnitude
falls from \DoseMagZero{} at $\alpha{=}0$ to \DoseMagOne{} at $\alpha{=}1$
(Spearman $\rho$ between dose and suppressed conflict: $\rhoDoseMan{}$).
The raw, unprojected level---also plotted---does \emph{not} fall with the
dose: the adapted gradients re-create conflict at comparable levels, so the
readout must come from the intervention itself. The causal readout (right):
does the trade-off follow? It does not---the Pareto score across doses is
\DoseClaim{}, with a dose--outcome correlation of $\rho = \rhoDoseCaus{}$
($p = \rhoDoseCausP{}$). Comparing full suppression ($\alpha{=}1$) with no
intervention ($\alpha{=}0$, identical to the matched E1 configurations) by
seed moves the joint score by only \alphaDelta{} (understanding
\alphaDeltaU{}, generation object-cell \alphaDeltaG{}), against a
seed-to-seed spread of \alphaSpread{}. Suppressing conflict completely
\AlphaClaim{}.

\subsection{E4: what does predict the trade-off?}
\label{sec:e4}
If gradient-geometric conflict metrics fail, the constructive question is
what succeeds. We compare four functional measures against the conflict
metrics at the checkpoint level (Table~\ref{tab:pred}; visual comparison in
Figure~\ref{fig:e4}): probe separability, subspace overlap, effective
rank, self-consistency (Section~\ref{sec:func}). The best functional
predictor (\bestAlt{}) reaches
$\rho = \predBestalt{}$ (CI \predBestaltLo{}--\predBestaltHi{}) versus
\predBestconfDir{} for the best directional conflict metric
(\bestConfDir{}), a difference of \BestClaim{} (paired bootstrap CI on
$\Delta\rho$: \deltaAltConfDirLo{}--\deltaAltConfDirHi{}). The strongest
gradient-geometric competitor is the norm ratio ($\rho = \predNormratio{}$,
CI \predNormratioLo{}--\predNormratioHi{}), comparable in absolute strength
($|\Delta\rho| = \absGapAltNr{}$, paired bootstrap CI
\absGapAltNrLo{}--\absGapAltNrHi{})---but the regime cut dissolves its
interest: as E1 established, the norm ratio is a generation-failure detector,
null within the configurations that master generation, where the functional
measures retain their advantage (\bestAltHealthy{} at
$\rho = \predBestaltHealthy{}$ versus \predBestconfHealthy{}, paired
$\Delta\rho$ CI \deltaAltConfHealthyLo{}--\deltaAltConfHealthyHi{}).
Self-consistency shows why the cut matters: the collapsed SOAP-style runs
are trivially self-consistent---they render uniform background scenes and
answer questions about them correctly---flipping the measure's sign in the
full population ($\rho = \predSelfconsistency$).

\begin{table}[t]
\centering
\caption{Predictor competition at the checkpoint level: Spearman $\rho$ with
the Pareto score over all \Nckpts{} checkpoint rows, cluster-bootstrap CIs.
Functional measures in the upper block, gradient-geometric conflict metrics
in the lower block.}
\label{tab:pred}
\resizebox{0.4\linewidth}{!}{%
\begin{tabular}{lrl}
\toprule
Predictor & $\rho$ & 95\% CI \\
\midrule
Probe separability & \predProbeacc{} & [\predProbeaccLo{}, \predProbeaccHi{}] \\
Subspace overlap & \predSubspaceoverlap{} & [\predSubspaceoverlapLo{}, \predSubspaceoverlapHi{}] \\
Effective rank & \predEffrank{} & [\predEffrankLo{}, \predEffrankHi{}] \\
Self-consistency & \predSelfconsistency{} & [\predSelfconsistencyLo{}, \predSelfconsistencyHi{}] \\
\midrule
Norm ratio & \predNormratio{} & [\predNormratioLo{}, \predNormratioHi{}] \\
Conflict magnitude & \predConflictmag{} & [\predConflictmagLo{}, \predConflictmagHi{}] \\
Removed energy & \predRemovedenergy{} & [\predRemovedenergyLo{}, \predRemovedenergyHi{}] \\
Conflict rate & \predConflictrate{} & [\predConflictrateLo{}, \predConflictrateHi{}] \\
Global cosine & \predCosglobal{} & [\predCosglobalLo{}, \predCosglobalHi{}] \\
\bottomrule
\end{tabular}}
\end{table}

\subsection{E5: does the conclusion survive a scale change?}
\label{sec:e5}
Replicating the audit at a larger scale ($d{=}\Ld$, \Llayers{} layers,
\nLarge{} runs of three representative methods) yields the same sign
structure: \ScaleClaim{} (per-metric bars for both scales in
Figure~\ref{fig:e5}).

\subsection{Reliability ceiling and statistical power}
\label{sec:reliab}
A correlation cannot exceed the reliability of its measures
\citep{spearman1904}, and an audit that ignores this can mistake measurement
noise for a null result. We resample the held-out sets five times per final
checkpoint and compute ICC(1,1) \citep{shrout1979}: \IccAccu{} for
understanding accuracy, \IccAccgObj{} for generation object-cell accuracy.
The geometric-mean reliability implies a correlation ceiling of
$\sqrt{\mathrm{rel}_u \cdot \mathrm{rel}_g} = \reliabCeiling{}$;
disattenuating the observed config-level correlation gives \rhoGlobalDisatt{}---the
association remains weak after crediting the metrics for all attenuation
benchmark noise can explain. On power: with $N = \nEarly{}$ configurations
the minimum detectable $|\rho|$ at $\alpha{=}0.05$, power $0.8$ is
\mdeConfig{}; the checkpoint level ($N = \Nckpts{}$) reaches \mdeCkpt{}.
The audit is therefore powered to detect moderate associations and the
reported CIs are informative, not vacuous.

\section{Discussion and Limitations}
\label{sec:discussion}

\paragraph{What the audit does and does not show.} The result is a statement about metric validity, not about the usefulness
of any intervention. In the controlled regime we can establish the premise's
operational content---that a conflict metric measured during training
carries information about the eventual trade-off---and find it unsupported
for the directional metrics the field keys on: the association is
\DiscussionStrength{}, and it does not survive the reliability ceiling
correction, the progress control, or a per-layer reweighting. The causal
intervention shows that even \emph{removing} conflict does not move the
trade-off, ruling out the strongest rescue ("the correlation is diluted but
causally real"). None of this contradicts the possibility that on
production-scale UMMs conflict metrics behave differently; it establishes
that the burden of demonstration sits with the claimant, and provides the
protocol that makes the demonstration routine rather than an act of faith.

\paragraph{Why the field adopted an unvalidated proxy.} Conflict metrics are
cheap, continuous, and measurable every step, whereas the trade-off requires
benchmark suites with known reliability problems \citep{recht2019,
musgrave2020,varoquaux2022}, and the proxy has an intuitive mechanism story.
Our results do not make the proxy useless; they show that its \emph{validity
is setting-dependent}, and that the settings where it was adopted never
checked it. The protocol of Section~\ref{sec:hyps} costs a fraction of the
intervention methods' compute and returns an explicit go/no-go signal.

\paragraph{The method population exhibits the pathologies the field fears.}
The method rows are themselves informative. Under an order-of-magnitude
task-norm imbalance, MGDA's minimum-norm combination collapses onto the
minority-gradient task: understanding stalls at \mMgdaU{} while generation
reaches \mMgdaObj{}---a pathology mirroring, in miniature, the
modality-imbalance complaints that motivated balancing methods
\citep{paretolora2026,ogmge2022,pmr2023}. SOAP-style preconditioning
collapses the generation axis outright (object-cell accuracy \mSoapObj{},
joint score \mSoapPareto{} in all nine runs)---and its conflict statistics
sit inside the healthy range (cosine \mSoapCos{}, rate \mSoapRate{}, versus
naive training's \mNaiveCos{} and \mNaiveRate{}): the worst method is
invisible to every directional conflict metric, and only the norm ratio
flags it, for the failure-detection reason established in E1. Method choice
changes \emph{where} a configuration lands on the two axes
(Table~\ref{tab:methods}) without any conflict metric tracking those
differences---precisely the decoupling the audit is designed to expose.

\paragraph{Functional measures won---carefully interpreted.} The best functional predictor (\bestAlt{}, $\rho = \predEffrank{}$)
outperformed every directional conflict metric by a clear paired-bootstrap
margin and retains its advantage in the regime that masters generation
(\bestAltHealthy{}, $\rho = \predBestaltHealthy{}$). The strongest
gradient-geometric competitor, the norm ratio ($\rho = \predNormratio{}$),
is a generation-failure detector, null among configurations that master
generation. The functional measures stay strong exactly where the field
would deploy them. We read this as evidence for the construct the field was
reaching for: interference between the tasks' \emph{computations},
manifesting in representation geometry and cross-task functional consistency
rather than parameter-space gradient directions. We do not claim these
measures are the answer; they are the leading candidates in a competition
that, to our knowledge, had never been run.

\paragraph{Limitations.} (i)~\emph{Scale}: the testbed is 3.2M--6.7M
parameters on a synthetic grammar; the audit's logic and protocol transfer to
production scales, its quantitative conclusions do not automatically; we
mitigate by reproducing the conflict phenomenology and releasing a protocol
whose cost scales linearly in the number of configurations.
(ii)~\emph{Regime}: generation is mastered by every configuration training
its gradient at natural scale (the token-normalized weighting collapses it in
a subset of seeds), so trade-off variation mostly lives on the understanding
axis; the audit's machinery is unchanged when both axes vary.
(iii)~\emph{Method coverage}: seven gradient-combination strategies spanning
the families used by the examined literature \citep{mtlsurvey2021};
architecture and data-scheduling interventions are not swept.
(iv)~\emph{Power}: with \Nconfigs{} configurations the minimum detectable
correlation is \mdeConfig{}; the CIs are the honest statement of what the
design can and cannot rule out, and the checkpoint-level analysis with
\Nckpts{} rows tightens detection to \mdeCkpt{}.
\section{Conclusion}
\label{sec:conclusion}

A family of UMM designs and optimizers is built on an assumption that has
never been tested: that lowering a gradient-conflict metric improves the
understanding--generation trade-off. We turned that assumption into four
falsifiable hypotheses, built a controlled testbed in which each has a
designated experiment, ran the audit across \Ntotal{} training
configurations with reliability ceilings and a causal intervention, and
found the assumption unsupported: directional conflict metrics carry
\ConclusionVerdict{} signal about the trade-off, suppressing conflict
causally does not improve it, and functional interference measures predict
it better---while training loss, in the same testbed, tracks the trade-off
strongly, locating the failure in gradient-conflict geometry specifically.

\clearpage
\subsection*{AI use statement}

In this work we used generative AI tools for the following tasks with
disclosure. AI tools were used to (i) implement the testbed code, training
protocol, measurement pipeline, and analysis scripts from a detailed written
research design; (ii) execute the training and evaluation runs on local
hardware and assemble the results into figures and tables; (iii) draft and
edit the manuscript text. We did not use AI tools for formulating the core
research hypotheses, which derive from the authors' literature audit
documented in the project's research-design files, and we did not use AI
tools to select or cite literature without verification: every citation was
checked against the arXiv or publisher record, and no results, statistics,
or bibliographic entries were generated without being traceable to an
executed computation or a verified record. All AI-assisted code was tested
for correctness---including end-to-end unit checks of data-to-target
alignment, teacher-forcing alignment, and cached-versus-full decoding
equivalence---and the audit's claims were read against the raw measurement
tables by the authors. We take full responsibility for the final content of
this work, including text, claims, and artifacts produced with the aid of
generative AI.

\subsection*{Reproducibility statement}

The complete audit is reproducible from the released artifact: the world and
model definitions, the seven gradient-combination methods and the
intervention, the fixed probe batch, all random seeds, the run manifest, and
the analysis script that regenerates every number, figure, and table in this
paper from raw measurement CSVs. Appendix~\ref{sec:app:hyper} lists all
hyperparameters; Appendix~\ref{sec:app:proto} documents the
teacher/student-forcing schedule and the per-method update rules, including
our two documented deviations from canonical implementations (norm
restoration for MGDA and RMS-normalized SOAP-style preconditioning). Each
run's configuration, hardware, wall-clock time, and software versions are
recorded in its manifest, and the full compute budget was a single Apple
M4 Max laptop (MPS) run with up to six concurrent training processes over
\CompHours{} aggregate run-hours of wall-clock time.

\subsection*{Ethics statement}

This work is a negative-result audit of a diagnostic practice in multimodal
model training. It involves no human subjects, no personal data, and no new
data collection; the testbed is fully synthetic. The main ethical exposure is
scholarly: negative results that contradict widely repeated premises can be
unwelcome to the works whose premises are examined, and we have therefore
restricted every claim to what the controlled setting demonstrates, cited
the examined works precisely and verifiably, and provided the protocol by
which any of our conclusions can be re-tested at other scales. We release
the audit protocol publicly so that practitioners with different compute
budgets can apply it to their own settings.

\bibliography{refs}
\bibliographystyle{iclr2027_conference}

\appendix
\section{Extended Related Work}
\label{sec:app:related}

\paragraph{Unified architectures, in detail.}
Show-o \citep{showo} unifies autoregressive and discrete-diffusion modeling
in one transformer; Emu3 \citep{emu3} demonstrates pure next-token
prediction over interleaved visual and textual tokens; Janus
\citep{wu2025janus} separates understanding- and generation-oriented visual
encoders while unifying the trunk, and Janus-Pro \citep{januspro2025} scales
that recipe with data and parameter scaling. Chameleon
\citep{chameleon2024} establishes early-fusion token-based mixed-modal
training with a stability recipe for image-text interleaves; Transfusion
\citep{transfusion2024} combines a language-modeling objective over text
with a diffusion objective over image patches in one model; Unified-IO~2
\citep{unifiedio2} extends autoregressive multimodality across vision,
language, audio, and action. Later unified systems pursue progressive
vocabulary learning \citep{ugen2025}, continuous-token autoregression
\citep{unifluid2025}, visual-autoregressive generation inside an MLLM
\citep{vargpt2025}, improved native unified modeling \citep{showo2},
visual-predictive instruction tuning \citep{metamorph2025}, native unified
image generation at scale \citep{hunyuanimage3}, and decoupled
autoregressive understanding-generation foundation models \citep{pisces2025};
surveys consolidate this rapidly growing family \citep{ummsurvey2025}.
Across all of these designs, one trunk (or one tied trunk family) serves
objectives with very different token budgets and output structures; our
testbed isolates exactly that shared structure.

\paragraph{Conflict-motivated UMM design, in detail.}
Uni-X \citep{hao2026unix} reports that gradient conflicts between vision and
text concentrate in shallow and deep layers of a shared autoregressive
transformer and diminish in middle layers, and proposes a two-end-separated,
middle-shared architecture; its evidence for downstream benefit is training
efficiency and benchmark scores, not a measured link between conflict
reduction and trade-off improvement. Symbiotic-MoE
\citep{liu2026symbioticmoe} frames understanding-side catastrophic
forgetting as the consequence of severe gradient conflicts and introduces
modality-aware expert disentanglement with early-stage gradient shielding.
ML-FOP-SOAP \citep{lu2026mlfopsoap} shows that first-order optimizers are
vulnerable to cross-modality gradient heterogeneity and proposes
Fisher-orthogonal projection with SOAP-style preconditioning
\citep{vyas2024soap}; its reported evidence is training-loss sample
efficiency and wall-clock speed on Janus and Emu3, without downstream
benchmark evaluation. Pareto LoRA \citep{paretolora2026} measures
order-of-magnitude modality-specific gradient imbalance under LoRA
fine-tuning and integrates text- and image-side gradients by Pareto
optimality. Task-aware MoE routing \citep{taskmoe2025} resolves
task-objective conflicts by routing.

\paragraph{Diagnosis without validation, in detail.}
\citet{rao2026dofight} perform a gradient analysis of DPO on Janus-Pro,
finding near-orthogonal understanding--generation gradients with 11--14$\times$
magnitude imbalance driven by visual token count asymmetry, and report that
no DPO variant improves generation CLIPScore ($p>0.5$); their analysis is
confined to preference tuning on a single architecture and does not measure
the conflict--trade-off correlation across a controlled population.
XTC-Bench \citep{wang2026crosstask} demonstrates that existing evaluation
treats understanding and generation independently and that high scores on
both do not imply cross-task consistency; UniEval similarly argues for
holistic evaluation of unified models \citep{unieval2025}.
\citet{wu2026synergy} study when the two objectives reinforce or compete in
controlled native settings and find one objective tends to dominate when
both are forced through the same computation path; \citet{su2026unigame}
exploit the understanding--generation inconsistency for post-training;
\citet{intrainter2025} study intra- and inter-modal forgetting in continual
UMM training. These works characterize the phenomenon or mitigate its
downstream symptoms; none validates the diagnostic metric itself.

\paragraph{Multi-task optimization, in detail.}
Joint training of multiple tasks predates deep multimodal models
\citep{caruana1997} and is surveyed for dense prediction
\citep{mtlsurvey2021}. Architectural sharing ranges from multi-gate mixture
of experts \citep{mmoe2018} to multi-scale task-interaction networks
\citep{mtinet2020}. On the optimization side, uncertainty-based loss
weighting \citep{kendall2018} and auxiliary-gradient-magnitude balancing
\citep{metabalance2022} adapt scalar weights; gradient manipulation rescales
or projects the vectors themselves: minimum-norm convex combinations
\citep{sener2018mgda}, projection-based gradient surgery and conflict-averse
cone constraints \citep[see][]{mtlsurvey2021,liu2021cagrad}, bargaining-game
Nash solutions \citep{nashmtl2022}, and independent-component alignment that
upweights conflict-prone gradient components \citep{senushkin2023}.
Pareto-dominance concepts enter multi-task evaluation from multi-objective
optimization \citep{deb2002}. Our method population spans these families at
faithful implementations, so that conflict metrics vary across methods
designed precisely to reshape conflict.

\paragraph{Modality imbalance, in detail.}
\citet{peng2020hard} show that end-to-end multimodal training can
underperform unimodal training because the gradient is dominated by the
faster-learning modality, and propose gradient modulation to rebalance it;
their on-the-fly modulation line continues with generalized-median
balancing \citep{ogmge2022} and intra-network modulation \citep{otfm2024}.
Prototypical rebalancing addresses the imbalance at the decision level
\citep{pmr2023}, instantaneous probe-and-rebalance methods monitor and
correct imbalance during training \citep{towardseq2025}, and
modality-laziness analyses document when the model ignores one modality
entirely \citep{scope2026}. These works establish that magnitude imbalance
between task gradients is real and consequential---the premise our
norm-ratio finding refines: in our testbed the imbalance statistic is
diagnostic of generation failure rather than of directional conflict, and is
null exactly where the field would deploy it.

\paragraph{Audits of benchmark validity, in detail.}
Recht et al.~\citep{recht2019} rebuilt ImageNet and CIFAR-10 test sets and
showed that classifier accuracy drops substantially, revealing overfitting
to reused evaluation data; follow-up work characterized the remaining
mistakes and their distribution \citep{vasudevan2022}. In metric learning,
\citet{musgrave2020} re-evaluated four years of claimed improvements under
a controlled protocol and found the ranking of methods largely unchanged.
Methodological reviews in medical imaging trace systematic failures to
unvalidated proxies and evaluation practices \citep{varoquaux2022}. Our
audit imports this discipline into UMM training diagnostics: the target
quantity (the trade-off) is measured without benchmark idiosyncrasy, the
predictor family (conflict metrics) is fixed before analysis, and the
conclusion is scoped by explicit reliability and power statements.

\section{Hyperparameters and Implementation Details}
\label{sec:app:hyper}

\begin{table}[h]
\centering
\caption{Training and evaluation hyperparameters. Base scale is the default;
the large scale (E5) differs only in the listed rows.}
\label{tab:hyper}
\resizebox{0.92\linewidth}{!}{%
\begin{tabular}{lll}
\toprule
 & Base & Large (E5) \\
\midrule
Transformer & $d{=}256$, \Nblocks{} blocks, 4 heads, FFN 512 &
$d{=}\Ld$, \Llayers{} blocks, FFN 640 \\
Parameters & 3.2M & 6.7M \\
Optimizer & AdamW ($\beta_1{=}0.9$, $\beta_2{=}0.999$, wd $0.01$) & same \\
Learning rate & \Nlr{}, warmup 100, cosine to $0.1\times$ & same \\
Batch & 64 (mixed tasks) & same \\
Steps & \Nsteps{} & 1080 \\
Student-forcing cap & \PfMax{} of samples, max 16/step & same \\
Data ratios & 1:2, 1:1, 2:1 (understanding:generation) & same \\
Seeds & $\{0,1,2\}$ (base), $\{0,1\}$ (large) & --- \\
Conflict snapshots & every 50 steps, fixed probe batch (32+32) & every 50 \\
Eval checkpoints & \EvalAt{} & 270/540/810/1080 \\
Held-out sets & 512 questions, 256 descriptions & same \\
Reliability resamples & 5 per final checkpoint & --- \\
Hardware & Apple M4 Max; MPS (up to 6 concurrent runs) & same \\
Precision & float32 & float32 \\
\bottomrule
\end{tabular}}
\end{table}

The vocabulary has 108 tokens: 8 grid symbols, 25 position markers, 25 cell
tokens, and 50 text tokens. Descriptions list 6--8 rectangular objects by
their cell tokens in randomized order; understanding questions are of three
types (cell color lookup, color count over the full scene, color-pair
adjacency). The fixed probe batch (seed 1234) and the fixed evaluation sets
(seed 777) are identical across all runs; reliability resamples use seeds
5101--5105; the student-forcing rollout generator is seeded per run (training
batch rollouts) and per snapshot with a run-independent seed (probe
measurements), so probe statistics are comparable across configurations.

\section{Method Implementations and Documented Deviations}
\label{sec:app:methods}

All methods receive per-group task gradients $g_u, g_g$ (embedding, each of
\Nblocks{} blocks, output head) from two independent backward passes and
produce the update written into the shared parameters. Method descriptions
follow the published algorithms and standard reviews
\citep{mtlsurvey2021,senushkin2023}.

\textbf{Naive joint.} $g_u + g_g$.

\textbf{Token-normalized weighting.} Loss weights
$w_t \propto T_t^{-1/2}$ normalized to sum to 2, where $T_u$ is the number
of answer tokens and $T_g$ the number of generation tokens in the batch;
this is the $\sqrt{\cdot}$-normalized variant of token balancing used in
multimodal joint training.

\textbf{GradNorm-style adaptive weighting} \citep[see][]{mtlsurvey2021}.
Loss weights $w_t$ are updated by gradient descent on the rate residual
$|r_t - \bar r_t|$, $r_t = w_t \lVert g_t\rVert / \lVert g_t^{(0)}\rVert$
with reference norms frozen after 100 steps, target rates
$\bar r_t = \mathcal{L}_t/\mathcal{L}_t^{(0)}$, using the closed form
$\partial \lVert w_t g_t\rVert/\partial w_t = \lVert g_t\rVert$; weights are
clamped to $[0.1, 10]$.

\textbf{PCGrad-style gradient surgery} \citep[see][]{mtlsurvey2021,
senushkin2023}. When $\langle g_u, g_g\rangle < 0$, each task gradient is
projected onto the other's normal plane and the projections are summed;
otherwise the naive sum. We use the original form without magnitude
renormalization.

\textbf{CAGrad} \citep{liu2021cagrad}. With
$g_0 = (g_u{+}g_g)/2$ and radius $c = \delta\lVert g_0\rVert$ at
$\delta{=}0.5$: each task's component orthogonal to $g_0$ is capped at
$c/K$ and added to $g_0$.

\textbf{MGDA-UB} \citep{sener2018mgda}. The two-task convex weights solve
$\min_{w\in[0,1]} \lVert w\,g_u + (1{-}w)\,g_g\rVert^2$ in closed form.
\emph{Documented deviation}: the convex combination is rescaled to the norm
of the average gradient. Without this, the update magnitude collapses as
gradients anti-align, confounding method comparisons at a fixed step budget;
norm restoration is the standard practice in MGDA implementations. Note
that restoration does not repair the direction pathology: when one task's
gradient norm dominates, the convex solution still allocates the update to
the minority direction; only the magnitude is restored.

\textbf{SOAP-style preconditioning} \citep{vyas2024soap}. For 2-D weight
matrices, Adam moments (with $\beta_1{=}0.9$, $\beta_2{=}0.98$, unlike the
AdamW runs) are maintained in the eigenbasis of EMA covariances
$GG^{\!\top}$ and $G^{\!\top}G$ with basis refresh every 50 steps and
bias-corrected moments; 1-D parameters use plain Adam; the update is
RMS-normalized per step. \emph{Documented deviation}: we omit SOAP's
momentum-translation scaling and apply RMS normalization instead, which
keeps the learning rate comparable across optimizer families; we refer to
this variant as SOAP-style preconditioning throughout.

\textbf{The $\alpha$-intervention (E3).} Per gradient group, the generative
gradient is replaced by
$g_g' = g_g - \alpha\,(g_g^{\!\top}\hat g_u)\,\hat g_u$ with
$\hat g_u = g_u/\lVert g_u\rVert$; the intervention is applied before the
method combination (with the naive combiner). Every group's task-gradient
cosine scales by $(1-\alpha)$ up to a norm correction, so the conflict level
is monotonically suppressed in $\alpha$; $\alpha{=}0$ reproduces the matched
E1 configuration exactly.

\section{Protocol Details}
\label{sec:app:proto}

\textbf{Teacher/student-forcing mixture.} Before step 200 the generation
loss is pure teacher forcing on the ground-truth color tokens at the marker
anchors; the student-forced share then ramps linearly to \PfMax{} by step
500 and stays there (capped at 16 of the batch's generation samples per
step). Student-forced rows render colors by sampling from the model at
temperature 1.0 (masked to the color alphabet) at the true marker anchors,
while the targets remain the ground-truth colors; the rendering loss is
computed at the marker positions and the end-of-stream position
(Section~\ref{sec:testbed}). Probe snapshots use the same mixture with a
run-independent generator seed per snapshot, making conflict statistics
comparable across configurations.

\textbf{Measurement alignment.} The audit required three correctness
invariants, each covered by a unit check in the released code
(\texttt{selfcheck.py}): (i)~the causal mask in every attention path
(full-sequence forwards and KV-cached prefill), verified by exact equality
(float64) of cached and full-forward greedy decoding over equal-length
batches, the calling convention used by evaluation; (ii)~the
grid-value-to-color convention, verified by a 500-scene round trip from
scene integers to tokens and back; (iii)~loss alignment, verified by
asserting that the batched rendering loss reproduces a brute-force per-row
computation in which color $g_j$ is read at the position of marker
$\langle p_j\rangle$ and the end-of-stream target at the last color-token
position, for rows of differing description lengths inside one padded
batch.

\textbf{Pareto score and frontier.} The score is the mean of the two raw
accuracies, $S = (\mathrm{acc}_u + \mathrm{acc}_{g})/2$; the empirical
Pareto frontier is computed on the raw two-dimensional accuracy plane over
all E1 configurations at the final checkpoint, and a configuration is
Pareto-dominated by another if the latter matches or exceeds it on both
axes with one strict inequality \citep{deb2002}. We deliberately avoid
min--max normalization: with a saturating generation axis it would amplify
resample noise into the score.

\textbf{Statistics.} Spearman rank correlations \citep{spearman1904};
percentile bootstrap confidence intervals with $B{=}5000$
\citep{efron1979} (pairs resampled at the configuration level, clusters
resampled at the checkpoint level so that all checkpoints of a run move
together); exact $p$-values from the permutation-free normal approximation
of the rank correlation under the null; partial Spearman by rank-regressing
out the step index. ICC(1,1) from five resamples \citep{shrout1979};
attenuation analysis via $\sqrt{\mathrm{rel}_u\,\mathrm{rel}_g}$; minimum
detectable effect from the Fisher-$z$ two-sample approximation at
$\alpha{=}0.05$, power $0.8$.

\section{Audit Design and Supplementary Experiment Material}
\label{sec:app:audit}

\begin{table}[h]
\centering
\caption{Audit design. All runs use the \gridumm{} testbed of
Section~\ref{sec:testbed}; hyperparameters in Appendix~\ref{sec:app:hyper}.}
\label{tab:setup}
\begin{tabular}{ll}
\toprule
Population (E1) & \Nmethods{} methods $\times$ \Nratio{} ratios $\times$
\Nseeds{} seeds $= \Nconfigs$ runs \\
Training budget & \Nsteps{} steps, batch 64, AdamW \Nlr{} \\
Conflict snapshots & every \ProbeEvery{} steps on a fixed probe batch \\
Checkpoints & steps \EvalAt{} (fraction of budget 0.25/0.5/0.75/1) \\
Interventions (E3) & $\alpha \in \{0.25, 0.5, 0.75, 1.0\}$, 3 seeds \\
Cross-scale (E5) & $d{=}\Ld$, \Llayers{} layers, \nLarge{} runs \\
Reliability & 5 independent held-out resamples per final checkpoint \\
\bottomrule
\end{tabular}
\end{table}

\begin{table}[h]
\centering
\caption{Conflict metrics vs.\ downstream Pareto score. $\rho_{\text{pred}}$:
early probe conflict against the final score ($N = \nEarly{}$
configurations); $\rho_{\text{conc}}$: both measured at the final checkpoint
($N = \Nconfigs$). Partial correlations control for training progress
(\Nckpts{} rows); CIs are percentile bootstrap, $B{=}5000$
\citep{efron1979}.}
\label{tab:cor}
\resizebox{0.98\linewidth}{!}{%
\begin{tabular}{lrrrrl}
\toprule
Metric & $\rho_{\text{pred}}$ & 95\% CI & $\rho_{\text{conc}}$ & 95\% CI &
partial $\rho$ (ckpt) \\
\midrule
Global cosine & \rhoPredCosglobal{} & [\rhoPredCosglobalLo{},
\rhoPredCosglobalHi{}] & \rhoCosglobal{} & [\rhoCosglobalLo{},
\rhoCosglobalHi{}] & \rhoCkCosglobalPartial{} \\
Conflict rate & \rhoPredConflictrate{} & [\rhoPredConflictrateLo{},
\rhoPredConflictrateHi{}] & \rhoConflictrate{} & [\rhoConflictrateLo{},
\rhoConflictrateHi{}] & \rhoCkConflictratePartial{} \\
Conflict magnitude & \rhoPredConflictmag{} &
[\rhoPredConflictmagLo{}, \rhoPredConflictmagHi{}] &
\rhoConflictmag{} & [\rhoConflictmagLo{}, \rhoConflictmagHi{}] &
\rhoCkConflictmagPartial{} \\
Norm ratio & \rhoPredNormratio{} & [\rhoPredNormratioLo{},
\rhoPredNormratioHi{}] & \rhoNormratio{} & [\rhoNormratioLo{},
\rhoNormratioHi{}] & \rhoCkNormratioPartial{} \\
Logged energy statistic & \rhoPredRemovedenergy{} &
[\rhoPredRemovedenergyLo{}, \rhoPredRemovedenergyHi{}] &
\rhoRemovedenergy{} & [\rhoRemovedenergyLo{}, \rhoRemovedenergyHi{}] &
\rhoCkRemovedenergyPartial{} \\
\midrule
Depth-weighted cosine & --- & --- &
\rhoCoswdepth{} & [\rhoCoswdepthLo{}, \rhoCoswdepthHi{}] & --- \\
Shallow-weighted cosine & --- & --- &
\rhoCoswshallow{} & [\rhoCoswshallowLo{}, \rhoCoswshallowHi{}] & --- \\
\bottomrule
\end{tabular}}
\end{table}

\begin{figure}[h]
\centering
\includegraphics[width=\linewidth]{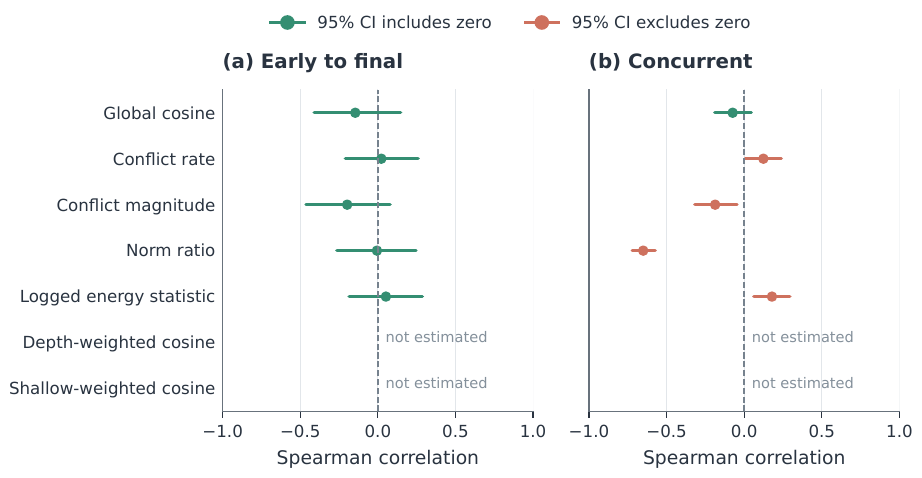}
\caption{Signed correlations with 95\% bootstrap intervals.
(a) Early probe geometry against final score, one observation per run.
(b) Concurrent association across all E1 checkpoints. Orange indicates
an interval excluding zero. Unavailable weighted estimates are marked
``not estimated'', not substituted from another analysis.}
\label{fig:corbars}
\end{figure}

\begin{figure}[h]
\centering
\includegraphics[width=\linewidth]{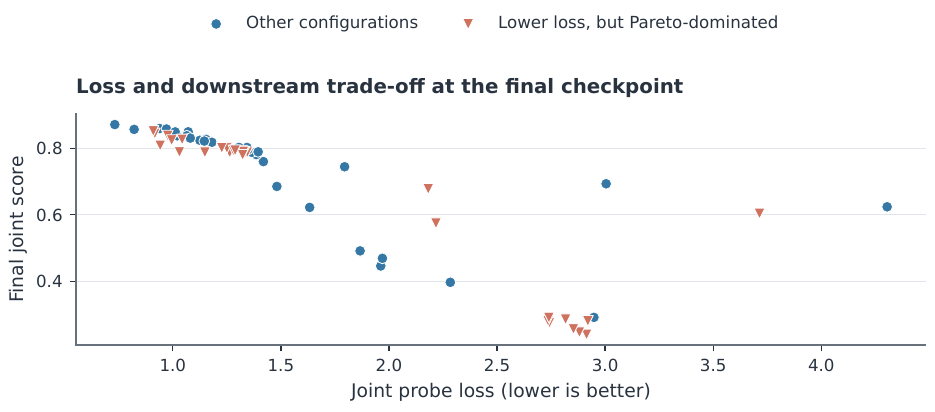}
\caption{Final joint loss vs.\ Pareto score. Red markers: configurations
whose training loss is better than another configuration's but whose
trade-off is Pareto-dominated by it (\domPairs{}/\domTotal{} pairs).}
\label{fig:e2}
\end{figure}

\begin{figure}[h]
\centering
\includegraphics[width=\linewidth]{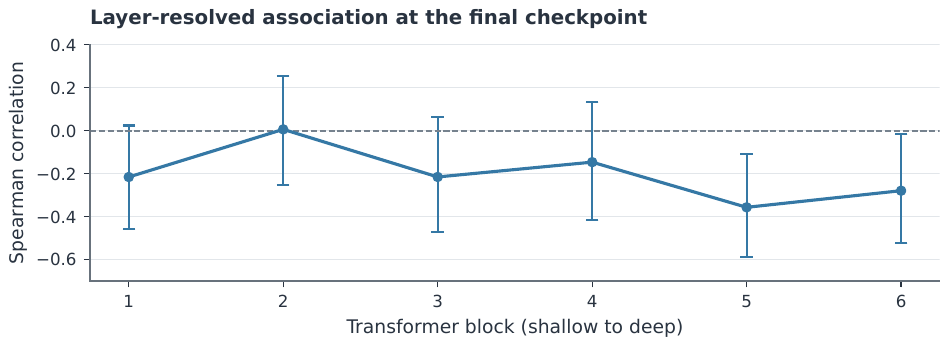}
\caption{Per-layer correlation profile: Spearman $\rho$ between each block's
task-gradient cosine and the Pareto score (configuration level, concurrent),
with 95\% CIs. The shallow-and-deep signature reported by Uni-X
\citep{hao2026unix} is not established by this concurrent profile:
blocks 5 and 6 have intervals excluding zero. The plot is an association
analysis, not a prospective layer-wise prediction test.}
\label{fig:layer}
\end{figure}

\begin{figure}[h]
\centering
\includegraphics[width=\linewidth]{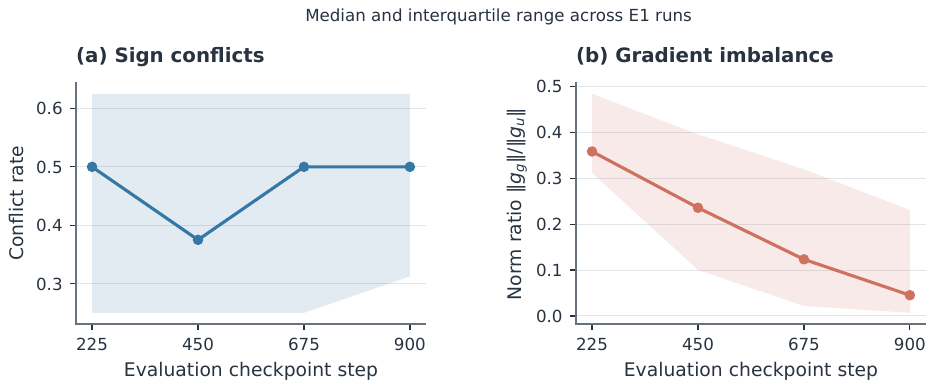}
\caption{The conflict regime across training: conflict rate (left) and
generation-to-understanding norm ratio (right), median and interquartile
range across all E1 configurations. The rate, a sign count over
near-zero cosines, stays elevated throughout; the norm ratio decays as the
generation objective converges.}
\label{fig:dynamics}
\end{figure}

\begin{figure}[h]
\centering
\includegraphics[width=\linewidth]{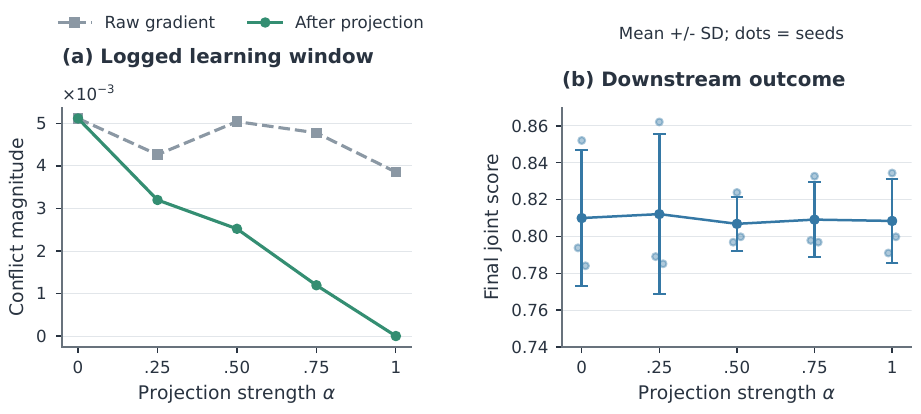}
\caption{Dose--response. Left: conflict magnitude under the $\alpha$-projection,
learning-phase mean over snapshots at steps 200--500 (the manipulation
check): the effective (post-intervention) level follows the dose, the raw
level shows co-adaptation. Right: the downstream Pareto score at the final
checkpoint (mean $\pm$ s.d.\ over seeds), the causal
readout.}
\label{fig:e3}
\end{figure}

\begin{figure}[h]
\centering
\includegraphics[width=\linewidth]{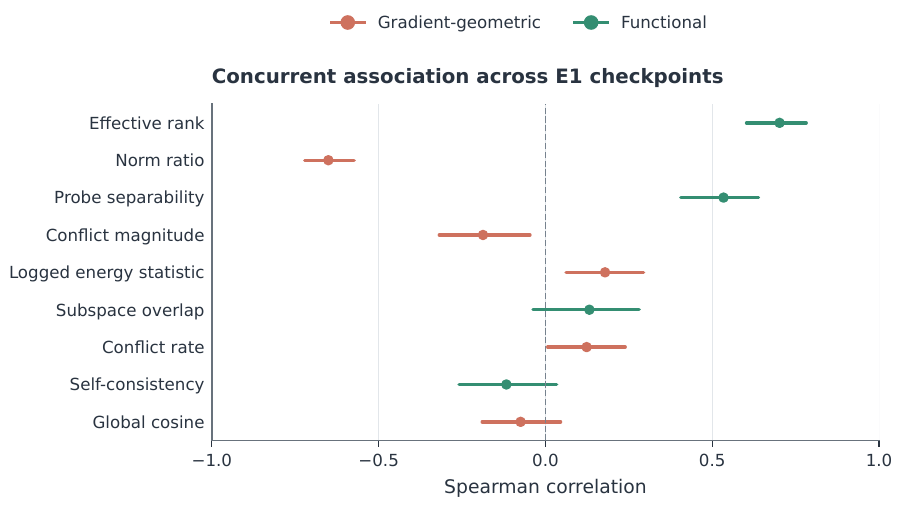}
\caption{Predictor competition at the checkpoint level: Spearman $\rho$
with the Pareto score, cluster-bootstrap CIs. Green: functional interference
measures; orange: gradient-geometric conflict metrics.}
\label{fig:e4}
\end{figure}

\begin{figure}[h]
\centering
\includegraphics[width=\linewidth]{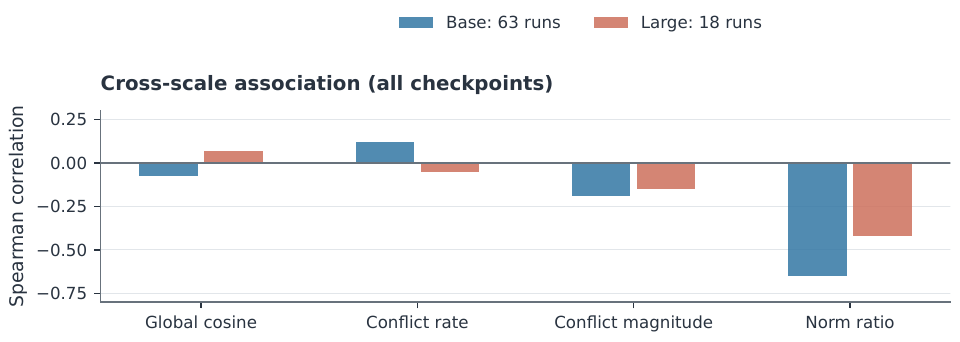}
\caption{Cross-scale check: per-metric correlations at the base and large
scales.}
\label{fig:e5}
\end{figure}

\begin{figure}[h]
\centering
\includegraphics[width=\linewidth]{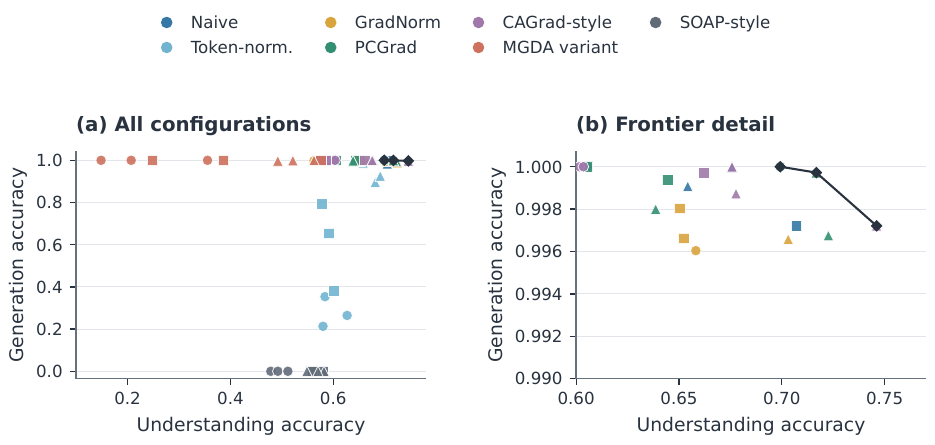}
\caption{The two-axis understanding/generation plane at the final checkpoint
over all E1 configurations, with the empirical Pareto frontier
(\frontierN{} configurations: \frontierMethods{}). Colors: methods; shapes:
data ratios.}
\label{fig:pareto}
\end{figure}

\textbf{Weighted and per-layer analyses (HA-4).} The separately stored,
early-concurrent weighted-cosine analysis referenced in E1 gives
\ConfClaimWtd{}; Table~\ref{tab:cor} lists the full conflict-metric vector
at both analysis levels, Figure~\ref{fig:corbars} shows the signed
correlation bars, and Figure~\ref{fig:layer} the per-layer profile that
contextualizes the layer-wise claims.

\textbf{Training dynamics.} Figure~\ref{fig:dynamics} documents the two
regime facts used in Section~\ref{sec:testbed}: the conflict rate stays
elevated throughout training while the generation-to-understanding norm
ratio decays as the generation objective converges, confining the learning
phase as the window in which the field's premise operates.

\textbf{Dose--response detail.} Figure~\ref{fig:e3} shows the E3
manipulation check and causal readout summarized in
Section~\ref{sec:e3}: the effective conflict level follows the dose, the
raw level shows co-adaptation, and the downstream Pareto score is flat
across doses.

\textbf{Predictor competition and cross-scale replication.}
Figure~\ref{fig:e4} visualizes the checkpoint-level predictor competition
summarized in Table~\ref{tab:pred}; Figure~\ref{fig:e5} shows the per-metric
correlations underlying the E5 statement: \ScaleClaim{}

\textbf{Pareto plane.} Figure~\ref{fig:pareto} plots the raw two-axis plane
whose empirical frontier contains \frontierN{} configurations
(\frontierMethods{}); method choice moves configurations along both axes
without any directional conflict metric tracking those differences
(Table~\ref{tab:methods}).

\end{document}